\documentclass[letterpaper, 10 pt, conference]{ieeeconf}  

\IEEEoverridecommandlockouts 

\usepackage{amsmath}
\usepackage{dsfont}
\usepackage{tabularx}
\usepackage{graphicx}
\usepackage[T1]{fontenc}
\usepackage[utf8]{inputenc}
\usepackage{babel}
\usepackage[raggedright]{titlesec}
\usepackage{blindtext}
\usepackage[dvipsnames]{xcolor}
\usepackage{xcolor,colortbl}
\usepackage{float}

\definecolor{Gray}{gray}{0.85}
\definecolor{LightCyan}{rgb}{0.88,1,1}
\renewcommand\footnotemark{}

\newcolumntype{a}{>{\columncolor{Gray}}c}

\definecolor{mypink2}{RGB}{2, 4, 143}

\title{\LARGE \bf Interpretable Long Term Waypoint-Based Trajectory Prediction Model}
\author{Amina Ghoul\textsuperscript{1},
 Itheri Yahiaoui\textsuperscript{2}, and Fawzi Nashashibi\textsuperscript{1}
}
\thanks{This work was carried out in the SAMBA collaborative project, co-funded by BpiFrance in the framework of the Investissement d’Avenir Program.\\
1: INRIA Paris, France {\tt firstname.lastname@inria.fr} ;\\
2: CReSTIC, Universit\'e de Reims Champagne-Ardenne, Reims, France 
{\tt itheri.yahiaoui@univ-reims.fr}}

\begin{document}
\maketitle
\thispagestyle{empty}
\pagestyle{empty}
\begin{abstract}
Predicting the future trajectories of dynamic agents in complex environments is crucial for a variety of applications, including autonomous driving, robotics, and human-computer interaction. It is a challenging task as the behavior of the agent is unknown and intrinsically multimodal. 
Our key insight is that the agents behaviors are influenced not only by their past trajectories and their interaction with their immediate environment but also largely with their long term waypoint (LTW). In this paper, we study the impact of adding a long-term goal on the performance of a trajectory prediction framework. We present an interpretable long term waypoint-driven prediction framework (WayDCM). WayDCM first predict an agent's intermediate goal (IG) by encoding his interactions with the environment as well as his LTW using a combination of a Discrete choice Model (DCM) and a Neural Network model (NN). Then, our model predicts the corresponding trajectories. 
This is in contrast to previous work which does not consider the ultimate intent of the agent to predict his trajectory.
We evaluate and show the effectiveness of our approach on the Waymo Open dataset.
\end{abstract}
\section{Introduction}

Predicting the future motion of a dynamic agent in an interactive environment is crucial in many fields and especially in autonomous driving.

A key challenge to future prediction is the high degree of uncertainty, in large part due to not knowing the behavior of the other agents. Because of these uncertainties, future motion of agents are inherently multimodal. Multimodality can be modeled by using implicit distributions from which samples can be drawn such as conditional variational
autoencoders (CVAEs) \cite{lee2017desire} or generative adversarial networks (GANs) \cite{gupta2018social}. The uncertainties can also be captured by the
prediction of possible intermediate goals of the agents  before predicting the full trajectories \cite{gilles2021gohome, gilles2021thomas, zhao2020tnt}. 

Moreover, most works predict the trajectories of moving agents by considering their past trajectories and their dynamic and/or static environment \cite{messaoud2020trajectory}.

However, we assume that before going anywhere, the agent knows its long-term waypoint, and its movement is therefore drawn to it. For example, we can consider the case of an agent entering his final destination on a GPS. The GPS then gives waypoints that need to be followed by the agent to reach its destination. In this paper, we study the impact of adding this waypoint information to our trajectory prediction model.

Furthermore, our model combines an interpretable discrete choice model with a neural network for the task of trajectory prediction. This combination allows for interpretable outputs. Our approach presents a way to easily validate NN models in safety critical applications, by using the interpretable pattern-based rules from the DCM.
\begin{figure}[!t]
\centering
\includegraphics[scale=0.5]{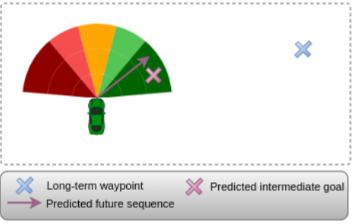}
\caption{The favorable directions are in green and the unfavorable directions in red. The cross in blue refers to the position of a long term waypoint of the vehicle. Knowing the position of this waypoint is an important cue for predicting its intermediate goal (the red cross). Here, the vehicle tends to go to right in order to get to reach the blue cross.}
\label{nn}
\vspace{-6mm}
\end{figure} 
\\
In this paper, we introduce \textbf{WayDCM}, which builds upon our last work TrajDCM \cite{ghoul:hal-04108657}, but taking into account a long term waypoint of the agents and proposing a first way to introduce it in a trajectory prediction framework. 
We conduct extensive experimentations on the real-world Waymo open dataset and we demonstrate the effectiveness of our method.

The content is organized as follows: Section II provides a
review on the background and current state of the research
fields. Section III introduces the method of the proposed
model. Section IV presents the different experiments
conducted. Section V presents the performance of the
approach for vehicle trajectory prediction and the comparison
to baseline models. Finally, a conclusion regarding the
architecture and results is drawn.
\section{Related Work}
\subsection{Input Representation}
In order to predict its trajectory, state of the art works use different forms of input cues that can reveal information about the future motion of the moving target agent.

The first main cue consists of the target agent's 
past state observations (positions, orientation, velocities, acceleration, etc.). Most studies use a sequence of past features to exploit the characteristics of the motion variation in the prediction task \cite{zhao2021tnt, deo2022multimodal}.

Then, the dynamic scene states represented by the past state observations of the surrounding agents are an other important cue, as their interaction with each other and with the target agent has an influence on the target agent's future motion.

Also, the static scene elements such as lanes and crosswalks, for example, help to determine the reachable
areas and therefore the possible patterns of the target agent motion. Different state of the art works exploit static scene
features and/or the dynamic ones \cite{ messaoud2021trajectory, gilles2021thomas} to infer the motion of the target agent.

In addition, some recent works such as \cite{zhong2022aware} consider the historical trajectories previously passing through a location as a new type of input cue in order to help infer the future trajectory of an agent currently at this
location. 

However, these approaches do not consider the long term waypoints that lead to the agent's final destination as an input cue. 

In our work, we propose a lightweight representation based on agents states and intermediate goals definition. We also add a long-term waypoint of the agent in order to better predict and understand the agent's behavior.
\subsection{Conditioning on Intermediate Goal}
Conditioning on intermediate goal (IG) enhances the ability of trajectory prediction models to capture the influence of explicit IGs on agent behavior and improve the accuracy of trajectory forecasts.
Several methods such as TNT \cite{zhao2021tnt} or LaneRCNN \cite{zeng2021lanercnn} condition each prediction on the intermediate goal of the driver. Conditioning predictions on future IGs helps leverage the HD map by restricting those IGs to be in a certain space.
MultiPath \cite{MultiPath} and CoverNet \cite{Covernet} chose to quantize the trajectories into anchors, where the trajectory prediction task is reformulated into anchor selection and offset regression. 
In this paper, we use a radial grid similar to \cite{ghoul:hal-04108657, kothari2021interpretable} to predict the intermediate goal of the moving agent.
\subsection{Interpretable Trajectory Prediction}
Most state-of-the-art studies use neural networks for the task of trajectory prediction. However, neural networks models are not interpretable. In fact, their outputs cannot be explained by humans.
In order to adress the lack of interpretability, recent studies focus on adding expert knowledge to deep learning models for trajectory prediction. A way to encourage interpretability in trajectory prediction architectures is through discrete modes. Brewitt et al. \cite{brewitt2021grit} propose a Goal Recognition
method where the “goal” is defined by many behavioral intentions, such as “straight-on”, “turn left” for example.
In \cite{yue2022human},  the authors propose a method combining a social force model, and
a method comprised of deep learning approaches based on a Neural Differential Equation model.  They use a deep neural network within which they use an explicit
physics model with learnable parameters, for the task of pedestrian trajectory prediction.
\cite{ghoul:hal-04108657} learn a probability distribution over possibilities in an interpretable discrete choice model (DCM) for the task of vehicle trajectory prediction. 

We use a similar approach as \cite{ghoul:hal-04108657} for predicting the trajectories of diverse agents. 
However, this paper differs from our previous work as we propose a method to include the long term waypoint information of the agent in a DCM model.
\subsection{Contributions}
To the best of our knowledge, we are the first to introduce the idea of a long term waypoint to a trajectory prediction framework. In fact, most studies do not take into account this information as they mostly model the interactions between agents \cite{alahi2016social, messaoud2020attention} or/and with the static environment \cite{messaoud2021trajectory, deo2022multimodal}. However, we argue that knowing where the agent is going, is a prior knowledge that should be considered in order to accurately predict its future trajectory. Indeed, while moving, an agent is not only influenced by its interactions with its environment, but also by its destination.

To summarize, we list the
main contributions of our approach as follows:
\begin{itemize}
    \item We introduce an interpretable DCM-based model which includes a long term waypoint of the agent, for trajectory prediction.
    \item We study the impact of adding such information on the predicted trajectories. 
    \item We implement and evaluate our approach on the Waymo Open Dataset. Our goal is to open the discussion about adding the long term waypoints of agents in motion prediction datasets. And also to propose a first approach that considers this new information.
\end{itemize}
\begin{figure*}[h]
\centering
\includegraphics[scale=0.53]{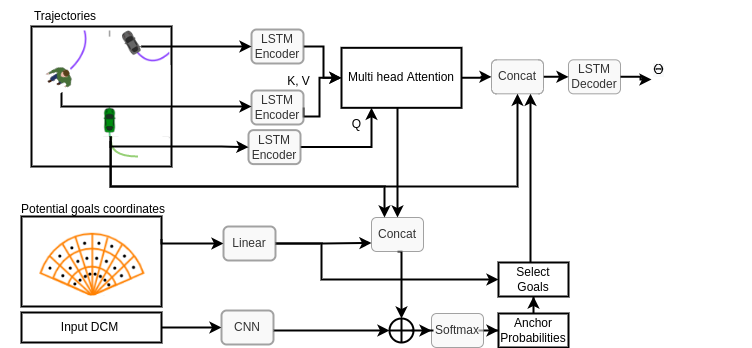}
\caption{Architecture of the method for trajectory prediction. The model take as inputs the past trajectories of the agents in the scene, the target coordinates sampled from a radial grid, as well as the input of the DCM model. It outputs $L$ trajectories. For more details see section \ref{model}.}

\vspace{-6mm}
\end{figure*}
\vspace{-2mm}
\section{Method}
\label{model}
\subsection{Problem definition}
The goal is to predict the future trajectories of a target agent $T$ : $\hat{Y_T}=(\hat{x}_T^t, \hat{y}_T^t)$ from time $t = t_{obs}+1$ to $t = t_f$.
We have as input of our model the track history of the target agent and the $n$ neighboring agents in a scene defined as 
$\textbf{X} = [X_1, X_2, ..., X_n]$. Each agent $i$ is represented by a sequence of its states, from time $t=1$ to $t=t_{obs}$. Each state is composed of a sequence of the agent relative coordinates $x_i^t$ and $y_i^t$, velocity $v_i^t$, heading $\theta_i^t$.
\begin{eqnarray}
X_i^t = (x_i^t, y_i^t, v_i^t, \theta_i^t)
\end{eqnarray}
The positions of each agent $i$ are expressed in a frame where the origin is the position of the target agent at $t_{obs}$. The x-axis is oriented toward the target agent’s direction of motion and y-axis points to the direction perpendicular to it.
\subsection{Utility function}
\label{dcm}
Similar to \cite{ghoul:hal-04108657}, we use a discrete choice model to help predict the intermediate goal of the agent among a discrete set of potential goals. The DCM allows to model the behavior of agents in their interactions with their surroundings.

The intermediate goals are extracted from a dynamic radial grid where the longitudinal size of the grid $maxl$ depends on the velocity of the target agent at $t_{obs}$. So that, we have $maxl = 1.5 \times v_T^{t_{obs}} \times t_f$.

We use the Random Utility Maximization (RUM) theory \cite{manski1977structure} that postulates that the decision-maker aims at maximizing the utility relative to their choice. The utility that an agent $i$ chooses an alternative $k$, is given as :
\begin{equation}
\begin{split}
u_k(\textbf{X}) & = \beta_{dir}dir_k + \beta_{occ}occ_k + \beta_{col}col_k \\ &+ \beta_{ddir}ddir_k + \beta_{ddist}ddist_k
\end{split}
\label{uk1}
\end{equation}
where $\beta_{dir}$, $\beta_{occ}$, $\beta_{col}$, $\beta_{ddir}$ and $\beta_{ddist}$  are the parameters associated with the explanatory variables $dir_k$, $occ_k$, $col_k$, $ddir_k$ and $ddist_k$ that describe the observed attributes of the choice alternative.

In the following, we describe each of these explanatory variables : 

\subsubsection{Keep direction $dir$}
This part of the model captures the tendency of people to avoid frequent variation of direction. Agents choose their position in order to minimize the angular displacement from their current direction of movement.

The variable $dir_k$ is defined as the angle in degrees between the direction of the alternative $k$ and the direction $dn$, corresponding to the
current direction of the agent.
\subsubsection{Occupancy $occ$}
We consider that alternatives containing neighbours in the vicinity are less desirable.
It is defined as the weighted number of agents being in the direction of the alternative $k$, that is :
\begin{equation}
    occ_k = \sum_{i=1}^{N}I_ke^{-dist_{ik}}
\end{equation}
where $N$ is the total number of agents in the environment, $I_k$ is one if the distance $dist_{ik}$ between the agent $i$ and the physical center of the alternative $k$ is less than $maxl / 3$.
\subsubsection{Collision avoidance $coll$}
The insight for this function is that when a neighbour agent’s trajectory is head-on towards a potential intermediate goal, this IG becomes less desirable due to the chance of a collision.

For each direction $d$, we choose a collider based on the following indicator function :

\begin{equation}
    I_{coll}^{i} =
    \begin{cases}
      1 & \text{if $d_l<d_i<d_r$, } \\
       & \text{and $0 < D_i < 2 \times max_l$,}\\
       & \text{and $\frac{\pi}{2}<|\theta_i - \theta_{k}|< \pi$},\\
      0 & \text{otherwise}
    \end{cases} 
\end{equation}
where $d_l$ and $d_r$ represent the bounding left and right directions of the cone in the choice set
while $d_i$ is the direction identifying the position of agent i. $D_i$ is the distance between agent $i$ and the target agent. $|\theta_i - \theta_{k}|$ is the difference between the movement direction of agent $i$, and the
direction of the alternative $k$.
Among the set of $K_d$ potential colliders for each direction, a collider $C$ is chosen in each cone as the agent that has the maximum value of $|\theta_i - \theta_{k}|$.
Finally, we have :
\begin{equation}
    coll_k = \alpha_{C} e^{\rho_{C} D_{C}},
\end{equation}
where $\alpha_{C}$ and $\rho_{C}$ are defined in \cite{antonini2006discrete}.

\subsubsection{Long term waypoint direction $dangle$}
We consider that an agent has the tendency  to choose, for the intermediate goal, a spatial location that minimizes the
angular displacement to the long term waypoint.

$dangle_k$ is defined as the angle in degrees between the destination and the long term waypoint of the alternative $k$.

\subsubsection{Long term waypoint distance $ddist$}
Here we consider that the agents tends to minimize the distance to the long term waypoint.
The variable $ddist_k$ is defined as the distance (in meters) between the long term waypoint and the center of the alternative $k$.

We expect a negative sign for all the $\beta$ parameters.

\subsection{Architecture}
For a target agent $T$ at time $t$, $X_T^t$ is embedded using a fully connected layer to a vector $e_i^t$ and encoded using an LSTM encoder,
\begin{equation}
h_i^t = LSTM(h_i^{t-1}, e_i^t; W_{enc}),
\end{equation}
$W_{enc}$ are the weights to be learned. The weights are shared between all agents in the scene.

Then we build a social tensor similar to \cite{messaoud2020trajectory}. 

We use the multi-head attention mechanism \cite{vaswani2017attention} to model the social interactions, where the target vehicle $h_T^{t_{obs}}$ is processed by a fully connected layer to give the query and the social tensor is processed by $1 \times 1$ convolutional layer to give the keys and the values. 

We consider $K$ attention heads where $K$ attention heads are specialized to the $K$ potential intermediate goals. 

For each attention head, we concatenate the output of the multi-head attention module $A_k$ with the target vehicle trajectory encoder state $h_T^{t_{obs}}$
to give a context representation $z_k$ for $k=1,... K$.
\begin{equation}
    z_k= Concat(h_T^{t_{obs}}, A_k)
\end{equation}
We then predict the intermediate goal by combining a DCM model with a NN model.

In order to help the knowledge-based model DCM capture the long term dependencies and the complex interactions, we use the Learning Multinomial Logit (L-MNL) \cite{sifringer2020enhancing} framework.\\
The IG selection probabilities is defined as :
\begin{equation}
\pi(a_k|\textbf{X}) = \frac{e^{s_k(\textbf{X})}}{\sum_{j \in K}e^{s_j(\textbf{X})}},
\end{equation}
where 
\begin{equation}
s_k(\textbf{X}) = u_k(\textbf{X}) + z_k(\textbf{X}),
\end{equation}

where $s_k(\textbf{X})$ represents the IG function containing the NN encoded terms, $z_k(\textbf{X})$, as well as utility function $u_k(\textbf{X})$, following the L-MNL framework.

The alternative $k$ corresponds to the target agent’s intermediate goal at timestep $t_f$, extracted from a radial grid, similar to \cite{ghoul:hal-04108657}.
We consider $L$ attention heads, for each attention head, we concatenate the output of the multi-head attention module $A_l$ with the target agent trajectory encoder state $h_T^{t_{obs}}$
to give a context representation $c_l$ for $l=1,... L$.
\begin{equation}
    c_l= Concat(h_T^{t_{obs}}, A_l)
\end{equation}
We select the $L$ best scored targets, and we concatenate their embedding to the output of the context representation $c_l$ for $l =1,... L$.

Finally, the context vector $c_l$ is fed to an LSTM Decoder which generates the predicted parameters of the distributions over the target vehicle’s estimated future positions of each possible trajectory for next $t_f$ time steps,
\begin{equation}
\Theta_l^t = \Lambda(LSTM(h_l^{t-1}, z_l; W_{dec})),
\end{equation}
where $W_{dec}$ are the weights to be learned, and $\Lambda$ is a fully connected layer.
Similar to \cite{ghoul:hal-04108657}, we also output the probability $P_l$ associated with each mixture component.

\subsection{Loss function}
Our proposed model outputs the means and variances $\Theta_l^t = (\mu_l^t, \Sigma_l^t)$ of the Gaussian distributions for each mixture component at each time step. \\
The loss for training the model is composed of a regression loss $L_{reg}$ and two classification losses $L_{score}$ and $L_{cls}$. $L_{reg}$ is the negative log-likelihood (NLL) similar to the one used in \cite{messaoud2020trajectory}, $L_{score}$ is a cross entropy loss and $L_{cls}$ is also a cross entropy loss defined as :
\begin{equation}
L_{reg} =  -\underset{l}{min}\sum\limits_{t=t_{obs} +1}^{t_{obs}+t_f}  log(\mathcal{N}(y^t|\mu_l^t; \Sigma_l^t))). 
\end{equation}
\begin{equation}
    L_{score} = -\sum_{l=1}^{L}\delta_{l*}(l)log(P_l), 
\end{equation}
where $\delta$ is a function equal to 1 if $l = l*$ and 0 otherwise.
\begin{equation}
    L_{cls} = -\sum_{k=1}^{K}\delta_{k*}(k)log(p_k), 
\end{equation}
where $p_k$ is the probability associated with the potential goal $k$, $\delta$ is a function equal to 1 if $k = k*$ and 0 otherwise, $k_*^t$ is the index of the potential intermediate goal most closely matching the endpoint of the ground truth trajectory. 

Finally, the loss is given by :
\begin{equation}
L = L_{cls} + L_{reg} + L_{score},
\end{equation}
\vspace{-5mm}
\begin{figure}
\end{figure}
\section{Experiments}
\subsection{Dataset}
Many datasets have been proposed for the task of motion prediction \cite{caesar2020nuscenes, zhan2019interaction, Argoverse, Argoverse2}. We choose to evaluate our approach on the Waymo Open dataset \cite{Ettinger_2021_ICCV} as it has a longer prediction horizon compared to the rest of the frequently used datasets (See Table. \ref{hor}). In fact, for the long term waypoint of the target agent, we consider the position at the longest time horizon.
Moreover, in the Waymo dataset, we can predict the future trajectories of agents for multiple horizons of 3, 5 or 8 seconds unlike the other datasets. We predict the future trajectories of a target agent for a horizon of $t_f = 3$ seconds while considering the long term waypoint at $8$ seconds. Therefore, we are able to compare our methods with other models only by using Waymo dataset.

The Waymo dataset provides 103,354, 20s 10Hz segments (over 20 million frames),  mined for interesting interactions. The data is collected from diferent cities in the United States of America (San Francisco, Montain View, Los Angeles, Detroit, Seattle and Phoenix).
\vspace{-5mm}
\begin{table}[h]

\begin{center}
\caption{Past, future and prediction horizons on different motion prediction datasets }
\label{hor}
\begin{tabular}{c c c c }
\hline
\rowcolor{Gray}
Dataset & Past (s) &  Future (s) & Prediction (s) \\
\hline
nuScenes & 3 & 5 & 5 \\ 
INTERACTION & 1 & 3 & 3\\ 
Argoverse 1 & 2 & 3 & 3 \\
Argoverse 2 & 5 & 6 & 6\\ 
\textbf{Waymo} & \textbf{1} & \textbf{8} & \textbf{3, 5 and 8}\\ 
\hline
\end{tabular} 
\end{center}
\vspace{-9mm}
\end{table}
\subsection{Implementation details}
We predict the future trajectory of the target agent $T$ for a horizon of $t_f=3$ seconds. We consider the final destination as the position of the target agent at $t=8$ seconds.
We use $K = 15$ number of potential intermediate goals. The radial grid representation is dynamic, i.e it depends on the current velocity of the target agent. The interaction space is 40 m ahead of the target vehicle, 10 m behind and 25 m
on each side. We consider the neighbors situated in the interaction space at $t_{obs}$. We use $L+K = 6+15$ parallel attention operations. We use a batch size of 16 and Adam optimizer. The model is implemented using PyTorch \cite{paszke2019pytorch}.
\subsection{Compared Methods}
The experiment includes a comparison of different models:
\vspace{-5mm}
\label{methods}
\begin{itemize}
    \item LSTM : An LSTM model with past trajectory of the agents as input.
    \item MHA-LSTM \cite{messaoud2020attention}: This multi-head attention-based model takes as inputs the past trajectories of the agents in the scene and outputs $L$ trajectories with their associated probabilities. We use $L=6$ attention heads.
    \item MotionCNN \cite{konev2022motioncnn} takes as input an image of the scene and uses a CNN followed by a fully connected layer to output multiple trajectories.
    \item MultiPath++ \cite{https://doi.org/10.48550/arxiv.2206.10041}:  takes as input the agent state's history and the road network and predicts a distribution of future behavior parameterized as a Gaussian Mixture Model (GMM).
    \item TrajDCM \cite{ghoul:hal-04108657} : To predict the IG of the target agent, this model combines a DCM and a neural network model. It then predicts the corresponding full trajectories. The DCM model is composed of the three first functions described in \ref{dcm} ($dir$, $occ$ and $coll$).
    \item WayDCM (1) : The proposed model without the function $ddist$ in the utility function.
    \item WayDCM (2) : The proposed model described in \ref{model}.
\end{itemize}
\section{Results}
\subsection{Evaluation metrics}
In order to evaluate the overall performance of our method, we use the following two error metrics. 
\begin{itemize}
    \item \textbf{Minimum Average Displacement Error over k ($minADE_k$)} : The average of pointwise L2 distances between the predicted trajectory and ground truth over the k most likely predictions.
    \item \textbf{Minimum Final Displacement Error over k ($minFDE_k$)} : The final displacement error (FDE) is the L2 distance between the final points of the prediction and ground truth. We take the minimum FDE over the k most likely predictions and average over all agents.
    
\end{itemize}
\subsection{Comparison of Methods}
We compare the methods described in Section \ref{methods}. We evaluated all the models on the validation set and we predicted the agent's future motion for $t_f=3$ seconds. We could not compare our approach with the models in the Waymo leaderboard as our method needs the position at $8$ seconds which is not given in the test set. 
Table \ref{comp_met} shows the obtained results.
We can see that  adding both the distance and the angle in the DCM model improve the original model TrajDCM.
Furthermore, our method is more lightweight than Multipath++ and MotionCNN as it does not take into account any map information. We implemented these two models using their code available online. We can see that we outperform MotionCNN and we obtain comparable results for $MinADE_6$ against Multipath++. 
\begin{table}[h]
\begin{center}
\caption{Comparison of different methods on the Waymo validation set (3 secs horizon) }
\begin{tabular}{c c c }
\hline
\rowcolor{Gray}
Model & \textbf{$MinADE_6$} &  \textbf{$MinFDE_6$} \\
\hline
MotionCNN  & 0.3365 & 0.6145  \\
MultiPath++ & 0.2692 & 0.4951  \\
\hline
LSTM & 0.4018 & 0.8029  \\ 
MHA-LSTM & 0.3141 & 0.7577  \\ 
TrajDCM & 0.3060 & 0.7201  \\ 

WayDCM (1) & 0.2779 & 0.6261  \\ 
\textbf{WayDCM (2)} & \textbf{0.2721} & \textbf{0.6037}  \\
\hline
\label{comp_met}
\end{tabular} 
\end{center}
\vspace{-7mm}
\end{table}
\subsection{Interpretable outputs}
\vspace{-4mm}
\begin{table}[h]
\begin{center}
\caption{Estimated parameters $\beta$ }
\begin{tabular}{c c c c c c}
\hline
\rowcolor{Gray}
Model & \textbf{$\beta_{dir}$} &  \textbf{$\beta_{col}$} & \textbf{$\beta_{occ}$} & \textbf{$\beta_{dangle}$} & \textbf{$\beta_{ddist}$} \\
TrajDCM & -2.70 & -0.07 &  -0.06   & - & - \\
WayDCM (1) & -3.48 & -0.09 &  -0.04  & -15.23 & -  \\
WayDCM (2) & -2.64 & -0.05 & -0.06  &  -10.83 &  -20.86 \\

\hline
\label{comp_beta}
\end{tabular} 
\end{center}
\vspace{-6mm}
\end{table}

The estimated parameters of the utility functions Eq. \ref{uk1}  are reported in Table. \ref{comp_beta}.
We can see that the all of the coefficients $\beta$ are negative. This is coherent with what we expected in Section \ref{dcm}. \\
Moreover, we can see that the functions $ddist$ and $dangle$ contribute the most to the prediction of the IG as their absolute values are higher than the absolute values of the other parameters for the model WayDCM(2). The function $ddist$ is the most significant. It means that
 agents tends to choose the IG which minimize the distance to the long term waypoint.
 
We can also notice that the functions $occ$ and $coll$ are not significant as their parameters for all the models are close to zero. This can be explained as a lack of interactions involving collisions for example. We plan to test our method on more interactive datasets in the future.
\section{Conclusion and future work}
The main scope of this paper was not to propose a model that outperforms the state-of-the art. Instead, we proposed a new approach to solve the trajectory prediction problem. We showed that using a long term waypoint can help improve the prediction performance.
We argue that long term waypoints should be included in motion prediction datasets as they are important input cues for the model. In fact, agents tend to move towards them. The position of these waypoints can be known before predicting the future trajectory of an agent, if given by a GPS for example. \\ 
We proposed a first simple approach that includes this information through a DCM. 
However, we hope that this paper will encourage researchers to consider the long term waypoints for the task of trajectory prediction. \\
For future work, we plan to try and compare other ways to include this information using different datasets, in a motion prediction framework.
\vspace{-1mm}
\addtolength{\textheight}{-1cm}   





\bibliographystyle{IEEEtran.bst} 
\vspace{-3mm}
\bibliography{refs}

\end{document}